\documentclass[10pt,twocolumn,letterpaper]{article}

\usepackage[cvprfinal]{cvpr} % To produce the REVIEW version
\usepackage[accsupp]{axessibility}
\usepackage[dvipsnames]{xcolor}
\usepackage{url}
\usepackage{comment}
\usepackage{amsmath,amssymb}
\usepackage{subcaption}
\usepackage{pgfplots}
\usepackage{tikz}
\usepackage{natbib}
\usetikzlibrary{decorations.text, arrows.meta, fit, positioning, automata, quotes}
\usepackage{colortbl}
\usepackage{epsfig}
\usepackage{multirow}
\usepackage{algorithm}
\usepackage{algpseudocode}
\usepackage[skip=0pt]{caption}
\usepackage{soul}
\usepackage[utf8]{inputenc}
\usepackage{listings}
\usepackage[T1]{fontenc}
\usepackage{booktabs}
\usepackage{amsfonts}
\usepackage{nicefrac}
\usepackage{microtype}
\usepackage{numprint}
\usepackage{siunitx}
\usepackage{etoolbox}
\usepackage{cancel}
\newcommand{\ubold}{\fontseries{b}\selectfont}
\robustify\ubold
\usepackage{bbold}
\usepackage{wrapfig}
\usepackage[skip=2pt,font=scriptsize,labelfont=scriptsize]{subcaption}
\usepackage{pifont}
\usepackage{lineno}
\usepackage{array,multirow,graphicx}
\usepackage[para]{footmisc}
\definecolor{cvprblue}{rgb}{0.21,0.49,0.74}

\usepackage[pagebackref=true,breaklinks=true,colorlinks,bookmarks=true,pagebackref,citecolor=cvprblue]{hyperref}
\usepackage{diagbox}

\usepackage{graphicx}
\pgfplotsset{compat=1.18}
\newcommand{\cmark}{\textcolor{OliveGreen}{\ding{51}}}
\newcommand{\xmark}{\textcolor{BrickRed}{\ding{55}}}
\usepackage[capitalize]{cleveref}
\crefname{section}{Sec.}{Secs.}
\Crefname{section}{Section}{Sections}
\Crefname{table}{Table}{Tables}
\crefname{table}{Tab.}{Tabs.}

\title{HIG: Hierarchical Interlacement Graph Approach to \\ Scene Graph Generation in Video Understanding}

\begin{document}

\author{Trong-Thuan Nguyen, Pha Nguyen, Khoa Luu\\
\small CVIU Lab, University of Arkansas \\
\tt\small\{thuann, panguyen, khoaluu\}@uark.edu \\
\small \tt \href{https://uark-cviu.github.io/ASPIRe/}{https://uark-cviu.github.io/ASPIRe/}
}

\twocolumn[{
 \renewcommand\twocolumn[1][]{#1}%
 \maketitle
 \begin{center}
 \centering
 % \vspace{-5mm}
 \captionsetup{type=figure}
 \includegraphics[width=.86\textwidth]{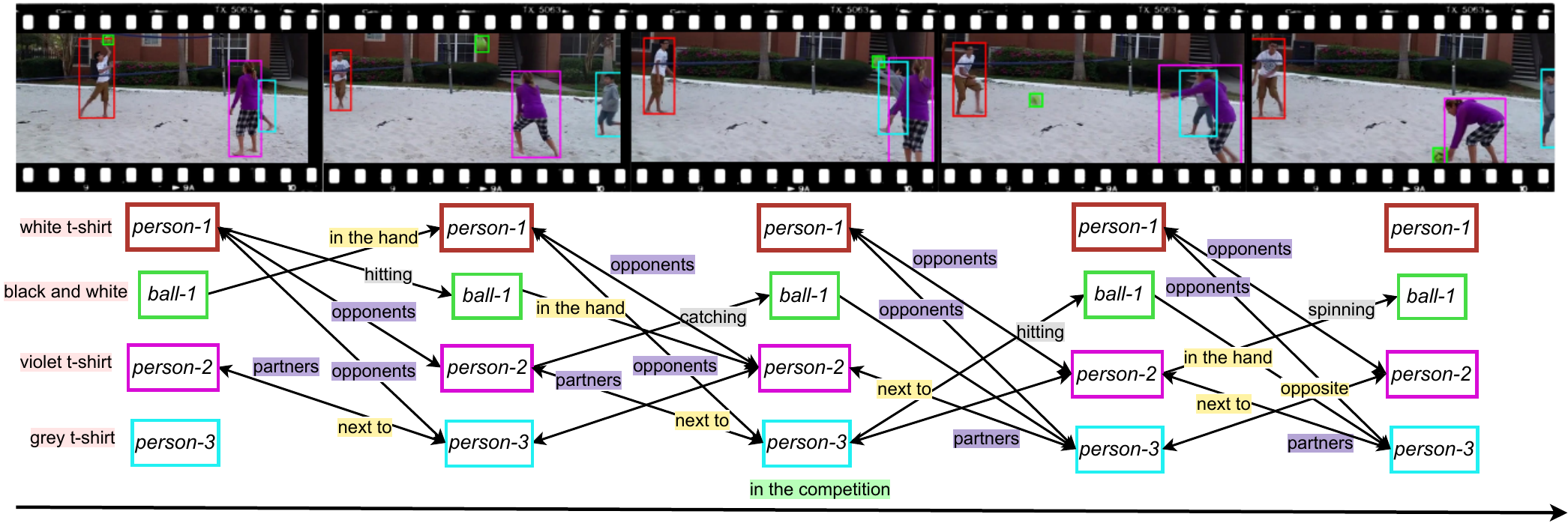}
 % \vspace{1mm}
 \captionof{figure}{An example from our \textit{ASPIRe} dataset for Visual Interactivity Understanding. The top row shows keyframes with the bounding boxes. \colorbox{pink!50}{\textbf{Appearance}},  \colorbox{green!30}{\textbf{Situation}}, \colorbox{yellow!50}{\textbf{Position}}, \colorbox{gray!25}{\textbf{Interaction}}, and \colorbox{blue!25}{\textbf{Relation}} are attributes presented in the dataset. \textbf{Best viewed in color}.
% The timeline tubes present temporal annotations. 
% \textcolor{red}{[@KL: Consider to compress the figure space. (Nov. 13) Consider to change another sample video.]} 
 }
 \label{fig:problem}
 \end{center}
 }]

\begin{abstract}

 %\textcolor{red}{[@KL: In Computer Vision, "Object" is understood differently from "human/subject". Our problem is modeling both "objects" and "subjects". So, "Agent Interactivity Understanding" is NOT A RIGHT TERM. Consider changing "Object" to "Agent"]The problem is change to Agent Interactivity Understanding}

%\textcolor{red}{[@KL: Let's try to complete the first draft by today, Friday. I will start revising the paper this weekend.] }

Visual interactivity understanding within visual scenes presents a significant challenge in computer vision. Existing methods focus on complex interactivities while leveraging a simple relationship model. These methods, however, struggle with a diversity of appearance, situation, position, interaction, and relation in videos. This limitation hinders the ability to fully comprehend the interplay within the complex visual dynamics of subjects. In this paper, we delve into interactivities understanding within visual content by deriving scene graph representations from dense interactivities among humans and objects. To achieve this goal, we first present a new dataset containing Appearance-Situation-Position-Interaction-Relation predicates, named \textit{ASPIRe}, offering an extensive collection of videos marked by a wide range of interactivities. Then, we propose a new approach named Hierarchical Interlacement Graph (HIG), which leverages a unified layer and graph within a hierarchical structure to provide deep insights into scene changes across five distinct tasks. Our approach demonstrates superior performance to other methods through extensive experiments conducted in various scenarios.

\end{abstract}

%%%%%%%%% BODY TEXT

\section{Introduction}

\begin{table*}[!t]
 \center
 % \scriptsize 
 \caption{Comparison of available datasets. \# denotes the number of the corresponding item.
 % \textbf{Bold} numbers are the best in each sub-block.
 % while \colorbox{red!25}{\textbf{highlighted}} numbers are the best across all sub-blocks.
 The top sub-block of the table is the summary of image datasets, and the bottom is video datasets.
 %The interactivity is the \textbf{appearance (\textit{A})}, \textbf{situation (\textit{S})}, \textbf{position (\textit{P})}, \textbf{interaction (\textit{I})}, and \textbf{relation (\textit{Re})}. 
 \textbf{Single} and \textbf{Double} are the attribute types as defined in Subsec.~\ref{subsec:terms}. \textbf{H-H}, \textbf{H-O}, \textbf{O-O} indicate the interactivity between \textit{Human and Human}, \textit{Human and Object}, \textit{Object and Object}.
 %\textcolor{red}{[@KL: This table must be cited in text. Also, This table has not emphasized the novelty (and challenge) of the problem. For example: Show relationship between object-object (Yes/No), object-human (Yes/No), etc.]} 
 % \textcolor{red}{[@KL: (Nov. 5, 23) Who takes care of this section? I don't see update for awhile!]The table includes more information, but we still need to consider the interactivity because the previous datasets didn't classify it into different types.}
 } \label{benchmarkList}
 \label{tab:datasets}
 \resizebox{\textwidth}{!}{
 \begin{tabular}{l|cccccccccccc}
 \toprule
 \multicolumn{1}{c}{\multirow{3}{*}{\textbf{Datasets}}} & \multicolumn{1}{c}{\multirow{3}{*}{\textbf{\#Videos}}} & \multicolumn{1}{c}{\multirow{3}{*}{\textbf{\#Frames}}} &
 \multicolumn{1}{c}{\multirow{3}{*}{\textbf{\#Subjects}}} &
 \multicolumn{1}{c}{\multirow{3}{*}{\textbf{\#RelCls}}} &
 \multicolumn{1}{c}{\multirow{3}{*}{\textbf{\#Settings}}} & %
 \multicolumn{3}{c}{\textbf{\multirow{2}{*}{\textbf{Annotations}}}} &
 \multicolumn{4}{c}{\textbf{Attributes}} \\
 \multicolumn{1}{c}{} & \multicolumn{1}{c}{} & \multicolumn{1}{c}{} & \multicolumn{1}{c}{} & \multicolumn{1}{c}{} & % <-- New column space
 \multicolumn{1}{c}{} & \multicolumn{1}{c}{} & \multicolumn{1}{c}{} & \multicolumn{1}{c}{} &
 \textbf{Single} & \multicolumn{3}{c}{\textbf{Double}} \\
 \multicolumn{1}{c}{} & \multicolumn{1}{c}{} & \multicolumn{1}{c}{} & \multicolumn{1}{c}{} & \multicolumn{1}{c}{} & \multicolumn{1}{c}{} & \textbf{BBox} & \textbf{Mask} & \multicolumn{1}{c}{\textbf{\#Annotations}} & \multicolumn{1}{c}{} & \textbf{H-H} & \textbf{H-O} & \textbf{O-O} \\
 \midrule
 \textbf{Visual Genome~\cite{krishna2017visual}} & - & \textbf{108K} & {\textbf{33K}} & {\textbf{42K}} & 1 & \cmark & \xmark & \textbf{3.8M} &
 \xmark & \xmark & \cmark & \cmark \\
 \textbf{PSG~\cite{yang2022panoptic}} & - & 49K & 80 & 56 & 1 & \cmark & \cmark & 538.2K &
 \xmark & \cmark & \cmark & \cmark \\
 \midrule
 \textbf{VidOR~\cite{shang2019annotating}} & {\textbf{10K}} & - & 80 & 50 & 1 & \cmark & \xmark & 50K &
 \xmark & \cmark & \cmark & \cmark \\
 \textbf{Action Genome~\cite{ji2020action}} & {\textbf{10K}} & 234K & 25 & 25 & 1 & \cmark & \xmark & \textbf{476.3K} &
 \xmark & \xmark & \cmark & \xmark \\
 \textbf{VidSTG~\cite{zhang2020does}} & {\textbf{10K}} & - & 80 & 50 & 1 & \cmark & \xmark & 50K &
 \xmark & \cmark & \cmark & \cmark \\
 \textbf{EPIC-KITCHENS~\cite{damen2022rescaling}} & 700 & 11.5K & 21 & 13 & 1 & \cmark & \xmark & 454.3K &
 \xmark & \xmark & \cmark & \xmark \\
 \textbf{PVSG~\cite{yang2023panoptic}} & 400 & 153K & 126 & 57 & 1 & \cmark & \cmark & - &
 \xmark & \cmark & \cmark & \cmark \\
 \textbf{ASPIRe (Ours)} & 1.5K & {\textbf{1.6M}} & \textbf{833} & \textbf{4.5K} & {\textbf{5}} & \cmark & \cmark & {167.8K} &
 \cmark & \cmark & \cmark & \cmark \\
 \bottomrule
 \end{tabular}
 }
 \vspace{-4mm}
\end{table*}

%\textcolor{red}{[What is "Interlacement"? Can we find a simpler term? Try to avoid confusion in wording for reviewers.]} Answered in Fig.~\ref{fig:terms}

Visual interaction and relationship understanding have witnessed significant advancements in computer vision in recent years. Various methods, including deep learning, have been introduced, particularly in achieving advanced comprehension of diverse relationships for a holistic visual understanding. Traditional methods span from action recognition and localization to intricate processes like video captioning~\cite{ye2022hierarchical, yang2023vid2seq, lin2023exploring}, spatio-temporal detection~\cite{zhao2022tuber, wu2023newsnet} and video grounding~\cite{li2023winner, tan2023hierarchical, lin2023collaborative}. However, these tasks often interpret visual temporal sequences in a constrained, uni-dimensional way. In addition, relation modeling techniques, including scene graph generation~\cite{ji2020action, yang2022panoptic, yang2023panoptic} and visual relationship detection~\cite{shang2019annotating, zhang2020does}, adhere to predefined relation categories, limiting the scope for discovering more diverse relationships.

Delving into the Visual Interactivity Understanding problem (Fig.~\ref{fig:problem})~\cite{shang2019annotating, ji2020action, yang2023panoptic}, we introduce a new dataset, characterized by 5$\times$ larger interactivity types, including \textbf{A}ppearance-\textbf{S}ituation-\textbf{P}osition-\textbf{I}nteraction-\textbf{Re}lation, named \textit{ASPIRe}. To this end, we introduce the Hierarchical Interlacement Graph (HIG), a novel approach to the Interactivity Understanding problem. The proposed HIG framework integrates the evolution of interactivities over time. It presents an intuitive modeling technique and lays the groundwork for enriched comprehension of visual activities and complex interactivities. HIG operates with a unique \textit{unified layer} at every level to jointly process interactivities. This strategy simplifies operations and eliminates the intricacies of multilayers. Instead of perceiving video content as a monolithic block, HIG models an input video with a \textit{hierarchical structure}, promoting a holistic grasp of object interplays. Each level delves into essence insights, leveraging the strengths of different levels to capture scene changes over time.
% Furthermore, HIG employs a \textit{unified graph} structure to address five crucial tasks, promoting a holistic grasp of object interplays.

% To this end, a Hierarchical Interlacement Graph (HIG) stands out as a unified layer, and hierarchical architecture addresses interactivity over a spectrum of abstraction levels, making it well-suited for diverse video sequences. In addition, HIG is a unified graph tailored to five tasks with its multi-level design. HIG highlights its scalability, effortlessly accommodating varied scenarios to ensure a proficient paradigm of intricate temporal interactivities.

% novelty:

% - unified layer

% - hierarchy structure

% - unified graph for 5 tasks

In addition, the proposed HIG framework promotes dynamic \textit{adaptability} and \textit{flexibility}, empowering the model to adjust its structure and functions to capture the interactivities throughout video sequences. This adaptability is further showcased as the HIG framework proficiently tackles five distinct tasks, demonstrating its extensive flexibility in decoding various interactivity nuances. 
% At every level, a unified layer and graph structure eliminates the complexities associated with multiple layers to ensure a comprehensive understanding of object interactivity from appearance and situation to position, interaction, and relations. 
The proposed HIG framework is not confined to specific tasks or domains, emphasizing its broad applicability and potential.
% selling points:

% - scalability

% - flexibility

% - generality

\textbf{The Contributions of this Work.} There are three main contributions to this work. First, we develop a new dataset named \textit{ASPIRe} for the Visual Interactivity Understanding problem, augmented with numerous predicate types to capture the complex interplay in the real world. Second, we propose the Hierarchical Interlacement Graph (HIG), standing out with its hierarchical graph structure and unified layer to ensure scalability and flexibility, comprehensively capturing intricate interactivities within video content. Finally, comprehensive experiments, including evaluating other methods on our \textit{APSIRe} dataset and HIG model on both video and image datasets,  we prove the advantages of the proposed approach that achieves State-of-the-Art (SOTA) results.

\section{Related Work}
\label{sec:related}

% \textcolor{red}{[@KL: Concise and Precise this section. It should be as short as possible. Save space for our own new materials, figures and tables.]}

\subsection{Dataset and Benchmarks}
% \vspace{-1mm}
\noindent \textbf{Dataset.} Action Genome~\cite{ji2020action} introduces a comprehensive video database with action and spatiotemporal scene graph annotations. VidOR~\cite{shang2019annotating} and EPIC-KITCHENS~\cite{damen2022rescaling} focus on object and relationship detection and egocentric action recognition. Ego4D~\cite{grauman2022ego4d}, VidSTG~\cite{zhang2020does}, and PVSG~\cite{yang2023panoptic} further enrich scene understanding and video scene graph resources. These datasets provide crucial benchmarks for evaluating scene understanding, detailed in Table~\ref{tab:datasets}.
% Within activity understanding, a diverse range of datasets has been introduced to specific applications, encompassing both images and videos. Visual Genome~\cite{krishna2017visual} dataset includes extensive images with comprehensive annotations, enhancing object detection and complicated relationship recognition tasks. The PSG~\cite{yang2022panoptic} dataset drives progress in panoptic scene graph generation, while the Action Genome~\cite{ji2020action} dataset serves as a benchmark for recognizing actions within video sequences. In addition, video datasets such as VidOR~\cite{shang2019annotating} and EPIC-KITCHENS~\cite{damen2022rescaling} involve an object, relationship detection, and egocentric vision-based action recognition. Additionally, Ego4D~\cite{grauman2022ego4d}, VidSTG~\cite{zhang2020does}, and PVSG~\cite{yang2023panoptic} datasets provide a valuable resource to scene understanding and video scene graphs. Collectively, these datasets serve as fundamental benchmarks in computer vision, facilitating comprehensive evaluations of scene understanding and image/video analysis, as summarized in Table~\ref{tab:datasets}.

% These datasets drive progress within activity understanding, promoting the development and evaluation of state-of-the-art models for object detection, action recognition, and scene understanding.

\begin{figure*}[!t]
 \centering
 \begin{minipage}{.670\textwidth}
 {\includegraphics[width=\textwidth]{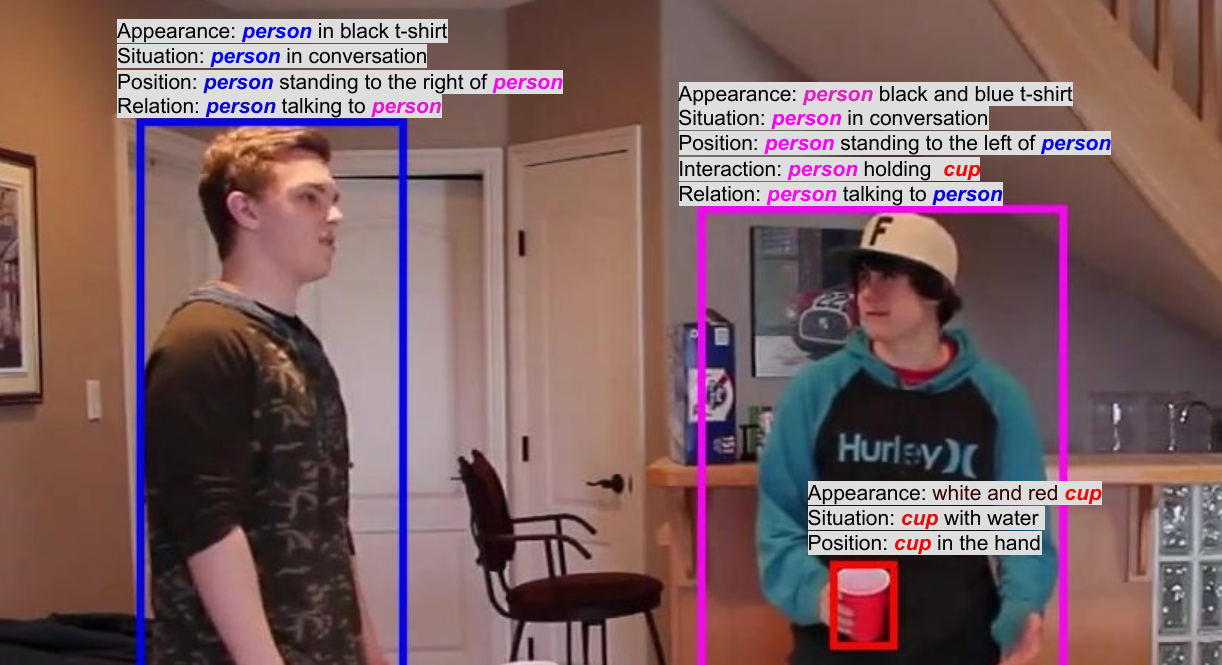}}
 \caption{Example and annotations in our \textit{ASPIRe} dataset. \textbf{Best viewed in color and zoom in}.}
 \label{fig:ex_annot}
 \end{minipage} \hfill
 \begin{minipage}{.3\textwidth}
 \subcaptionbox{\scriptsize A graph representation of the attributes in Fig.~\ref{fig:ex_annot}.}[\textwidth]{
 \resizebox{.75\textwidth}{!}{
\begin{tikzpicture}
 [
 > = stealth,
 shorten > = 1pt,
 auto,
 node distance = 3cm,
 semithick
 ]
 \tikzstyle{every state}=[
 draw = black,
 thick,
 fill = white,
 minimum size = 4mm
 ]
 \node[shape=circle,draw=black,fill=blue!50] (1) at (3,-2) {\textbf{person}};
 \node[shape=circle,draw=black,fill=magenta!50] (2) at (10,-3) {\textbf{person}};
 \node[shape=circle,draw=black,fill=red!50] (3) at (8,-6) {\textbf{cup}};

 \path[->, every node/.style={sloped,anchor=south,auto=false}] (1) edge [bend left] node[align=center] {standing to the right\\talking to} (2);
 \path[->, every node/.style={sloped,anchor=north,auto=false}] (2) edge [bend left] node[align=center] {standing to the left\\talking to} (1);
 \path[->, every node/.style={sloped,anchor=north,auto=false}] (2) edge [bend left] node {holding} (3);
 \path[->, every node/.style={sloped,anchor=north,auto=false}] (3) edge [bend left] node[align=center] {belonging to} (2);
 \path[->] (1) edge [align=center, loop above, thick, "black t-shirt\\conversation"] (1);

 \path[->] (2) edge [align=center, loop above, thick, "black and blue t-shirt\\conversation"] (2);

 \path[->] (3) edge [align=center, loop below, thick, "white and red\\in the hand\\with water"] (3);

 \end{tikzpicture}}}
 
 \subcaptionbox{\scriptsize Summary of annotated \textit{double-actor} attributes between two actors in our \textit{ASPIRe} dataset. \textit{appearance} and \textit{situation} are \textit{single-actor} attributes as in~\ref{subsec:terms}. \label{tab:interactivities}}
 [\textwidth]{\resizebox{.7\textwidth}{!}{\begin{tabular}{|ll|c|c|}
 \hline
 \multicolumn{2}{|l|}{\diagbox[height=1\line, width=7em]{\scriptsize $S_j$}{\scriptsize $S_i$}} & \textbf{\scriptsize Person} & \textbf{\scriptsize Object} \\ \hline
 \multicolumn{1}{|l|}{\multirow{3}{*}{{\rotatebox[origin=c]{90}{\scriptsize \textbf{Person}}}}} & \scriptsize Position & \cmark & \cmark \\ \cline{2-4}
 \multicolumn{1}{|l|}{} & \scriptsize Interaction & \xmark & \xmark \\ \cline{2-4}
 \multicolumn{1}{|l|}{} & \scriptsize Relation & \cmark & \cmark \\ \hline
 \multicolumn{1}{|l|}{\multirow{3}{*}{{\rotatebox[origin=c]{90}{\scriptsize \textbf{Object}}}}} & \scriptsize Position & \xmark & \xmark \\ \cline{2-4}
 \multicolumn{1}{|l|}{} & \scriptsize Interaction & \cmark & \xmark \\ \cline{2-4}
 \multicolumn{1}{|l|}{} & \scriptsize Relation & \xmark & \xmark \\ \hline
 \end{tabular}%
 }}
 \end{minipage}
 
 \vspace{-6mm}
\end{figure*}
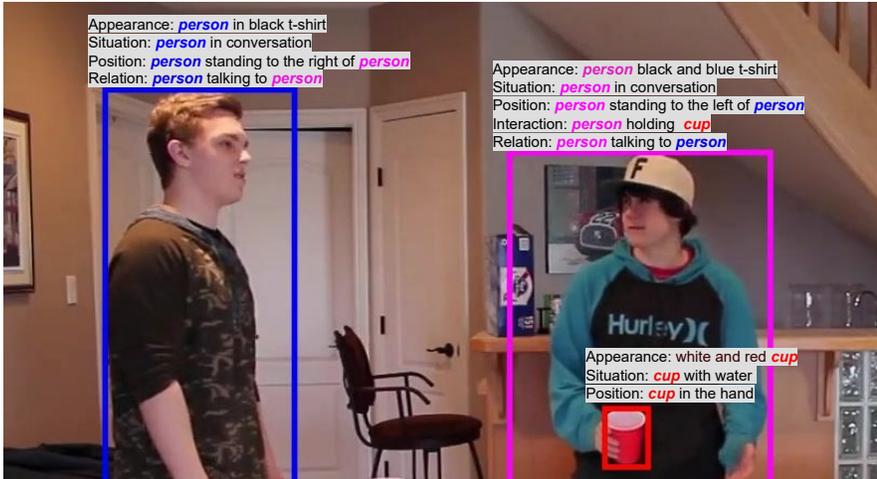

\noindent \textbf{Benchmarks.}
Current benchmarks primarily rely on relation classification for identifying inter-object associations.
% Previously, the Visual Genome~\cite{krishna2017visual} and PSG~\cite{yang2022panoptic} datasets played a crucial role in benchmarking scene graph generation. Specifically, PSG introduced PSG baselines to evaluate the Panoptic Segmentation Generation (PSG) task. While both datasets emphasized spatial information, encompassing many objects and categories of relationships in each frame, they needed to be more temporal information for capturing long-range dependencies in natural scenes.
Action Genome~\cite{ji2020action} integrates spatiotemporal to Visual Genome~\cite{krishna2017visual} to establish scene graphs with action recognition using SGFB. VidOR~\cite{shang2019annotating} provides 10K videos for benchmarking video object detection and visual relation detection. EPIC-KITCHENS-100~\cite{damen2022rescaling} offers a varied dataset with 100 hours of video, 20M frames, and 90K actions. Ego4D~\cite{grauman2022ego4d} focuses on first-person video data, addressing past, present, and future aspects across nearly 3.6K videos. VidSTG~\cite{zhang2020does} introduces the Video Grounding for Multi-Form Sentences (STVG) task, augmenting VidOR with additional sentence annotations. Recently, PVSG~\cite{yang2023panoptic} expanded PSG~\cite{yang2022panoptic}, advancing video graph generation.
% In contrast, Action Genome~\cite{ji2020action} is the first large-scale video database providing action and spatio-temporal scene graph labels. This benchmark incorporated Scene Graph Feature Banks (SGFB) to integrate spatiotemporal scene graphs into the action recognition task. VidOR~\cite{shang2019annotating} features 10K videos benchmarking for Video Object Detection and Visual Relation Detection. EPIC-KITCHENS-100~\cite{damen2022rescaling} introduces a dataset consisting of 100 hours of video, 20M frames, and 90K actions across 700 variable-length videos designed for various challenges related to understanding activities. Ego4D~\cite{grauman2022ego4d} focuses on first-person video data, addressing the past, present, and future aspects in nearly 3.6K video. VidSTG~\cite{zhang2020does} introduces a benchmark for the Video Grounding for Multi-Form Sentences (STVG) task, utilizing a large-scale video grounding dataset by augmenting sentence annotations on VidOR. Recently, Panoptic Video Segmentation (PVSG)~\cite{yang2023panoptic} is extended from PSG~\cite{yang2022panoptic} to capture temporal information.

\subsection{Interactivity Modeling Approaches}

\noindent \textbf{Video Situation Recognition.}
The VidSitu~\cite{sadhu2021visual} benchmark provides a collection of events and situations for evaluation, covering verb prediction, semantic role prediction, and event relations prediction. In a related approach within this benchmark, VideoWhisperer~\cite{khan2022grounded} adopts a global perspective for video comprehension, utilizing self-attention across all video clips. Furthermore, the LVU~\cite{xiao2022hierarchical} benchmark is tailored for self-supervised video representation learning, with a strong focus on hierarchical methodologies.

\noindent \textbf{Video Understanding.} This contains a wide range of tasks and research efforts. Action recognition~\cite{truong2022direcformer, quach2022non} has advanced significantly through graph-based~\cite{xing2023boosting}, few-shot learning~\cite{wang2022hybrid, thatipelli2022spatio}, and transformer-based~\cite{chen2022mm} approaches. Another area of interest is object retrieval~\cite{yang2023relational, moon2023query}, object tracking~\cite{nguyen2022multi, nguyen2024type}, spatio-temporal detection~\cite{liu2020beyond, zhao2022tuber, wu2023newsnet}, temporal audio-visual relationships~\cite{truong2021right} which involves object detection/segmentation, relation detection and moment retrieval in video content. Additionally, there are challenges such as visual question answering~\cite{xiao2022video, xiao2022video, urooj2023learning} and video captioning~\cite{ye2022hierarchical, yang2023vid2seq, lin2023exploring}. Recently, video grounding~\cite{yang2022tubedetr, li2023winner, tan2023hierarchical, lin2023collaborative} has provided activities through natural language in visual content.
% It facilitates the association of textual descriptions with specific moments in videos.

\noindent \textbf{Scene Graph Generation.} Biswas et al.~\cite{biswas2023probabilistic} introduce a Bayesian strategy for debiasing scene graphs in images, enhancing recall without retraining. PE-Net~\cite{zheng2023prototype} leveraging prototype alignment to improve entity-predicate matching in a unified embedding space, incorporating novel learning and regularization to reduce semantic ambiguity. PSGTR~\cite{yang2022panoptic} and PSGFormer~\cite{yang2022panoptic} introduce recent innovations in scene graph generation, which utilizes a transformer encoder-decoder to implicitly model scene graph triplets. Recently, PSG4DFormer~\cite{yang20234d} has been proposed to predict segmentation masks and then track them to create associated scene graphs through a relational component.
% , while PSGFormer adopts an explicit relation modeling approach with a "prompting-like" mechanism for precise subject-object-relation query matching.

For dynamic scenes, TEMPURA~\cite{nag2023unbiased} utilizes temporal consistency and memory-guided training to enhance the detection of infrequent visual relationships in videos. Cho et al.~\cite{cho2023davidsonian} introduce the Davidsonian Scene Graph (DSG) for assessing text-to-image alignment, operating a VQA module to process atomic propositions from text prompts and quantifying the alignment between text and image. Further, advancements by ~\cite{yang2022panoptic, goel2022not, li2022devil, lin2022hl} have adapted scene graph techniques to video, focusing on temporal relationships and advancing comprehensive scene understanding.

%\subsection{Discussion}
\subsection{Limitations of Prior Datasets}
% \vspace{-1mm}

Existing datasets exhibit notable limitations that hinder a comprehensive understanding of interactivity within visual content. Many of these datasets primarily focus on \textit{a limited set of interactivity types}, overlooking the complexity of real-world interactions. This restricted scope has impeded the development of models capable of handling a variety of interactivities, thereby limiting their applicability to diverse scenarios. Moreover, previous datasets predominantly emphasize relationships within \textit{single connected components of the relational graph}, neglecting complex scenes. Sparse annotations in some datasets further constrain relationship modeling, often failing to provide comprehensive coverage and potentially leading to model bias.
% Additionally, there is a substantial gap in the representation of scenes with dense crowds of people, which poses challenges in occlusion handling and complex interaction understanding.

To address these limitations, we introduce the new \textit{ASPIRe} dataset to Visual Interactivity Understanding. The diversity of the \textit{ASPIRe} dataset is showcased through its wide range of scenes and settings, distributed in seven scenarios. Therefore, \textit{ASPIRe} distinguishes itself from earlier datasets, including five types of interactivity, as in Fig.~\ref{fig:ex_annot}.
% In addition, it places a particular emphasis on scenarios with dense populations, thereby extending the frontiers of understanding complex interactivities.

\section{Dataset Overview}
% \begin{figure*}[!t]
% \centering
% \includegraphics[width=\textwidth]{figures/00001141.jpg}
% \caption{Example and annotations in our dataset. \textbf{Best viewed in color and zoom in}.} 
% \label{fig:egdata}
% \end{figure*}
\subsection{Dataset Collection and Annotation}
% \vspace{-1mm}
%\textcolor{red}{[@KL: This is the novelty (prior datasets don't have) and also one of main contributions of our work. Thus, spend time to present it. Structure this section so that it can highlight the importance and novelty of the proposed dataset.]}

We introduce a dataset compiled from seven distinct sources, each contributing unique perspectives to our collection. The ArgoVerse~\cite{chang2019argoverse} and BDD~\cite{bdd100k} datasets focus on outdoor driving scenes, providing valuable insights into real-world traffic scenarios. In contrast, the LaSOT~\cite{fan2019lasot} and YFCC100M~\cite{thomee2016yfcc100m} datasets consist of in-the-wild videos, capturing a diverse spectrum of human experiences and online interactions. Additionally, our dataset incorporates content from the AVA~\cite{gu2018ava}, Charades~\cite{charades}, and HACS~\cite{zhao2019hacs} datasets, encompassing videos that depict various human interactions, including interactions between humans and objects. This compilation results in a diverse scene featuring 833 objects. Therefore, the \textit{ASPIRe} dataset enhances the understanding of activities, surpassing traditional image datasets like Visual Genome~\cite{krishna2017visual} and PSG~\cite{yang2022panoptic} by integrating video data. This crucial integration brings a dynamic dimension to scene analysis that is conspicuously absent in static datasets. \textit{ASPIRe} stands out for its exceptional detail, demonstrating the dynamic interactivities over time. \textit{ASPIRe} has a depth of interactivities context that is notably comprehensive of other datasets while only presenting the relationship of humans, including VidOR~\cite{shang2019annotating}, Action Genome~\cite{ji2020action} and PVSG~\cite{yang2023panoptic}, marking a considerable stride in the scene understanding.

To this end, we introduce a structured annotation file anchored by a primary key named \texttt{data}. This file assembles dictionaries associated with a particular frame and detailed annotations. Each dictionary contains two crucial lists: \texttt{segments\_info} and \texttt{annotations}. The \texttt{segments\_info} list is a collection of dictionaries that describe the individual segments of the image, and the \texttt{annotations} list consists of dictionaries that offer bounding boxes and masking details for each segment. Additionally, objects identified within these segments and annotations are assigned the \texttt{track\_id} to maintain the identity within a video. In particular, the annotations within the \textit{ASPIRe} dataset are distinguished by five interactivity descriptors:
% \begin{itemize}
 (i) \texttt{appearances} details visual traits of subjects or objects;
 (ii) \texttt{situations} describes the environmental context;
 (iii) \texttt{positions} identifies the location and orientation;
 (iv) \texttt{interactions} captures the dynamic actions between \textit{Human-Object};
 (v) \texttt{relations} define the connections and associations between \textit{Human-Human}.
% \end{itemize}

\subsection{Dataset Statistics}
% \vspace{-1mm}
\begin{figure}[!t]
 \centering
 \begin{minipage}{.235\textwidth}
 \subcaptionbox{Video sources. \label{fig:stat_video}}[1.0\textwidth]{\includegraphics[width=\textwidth]{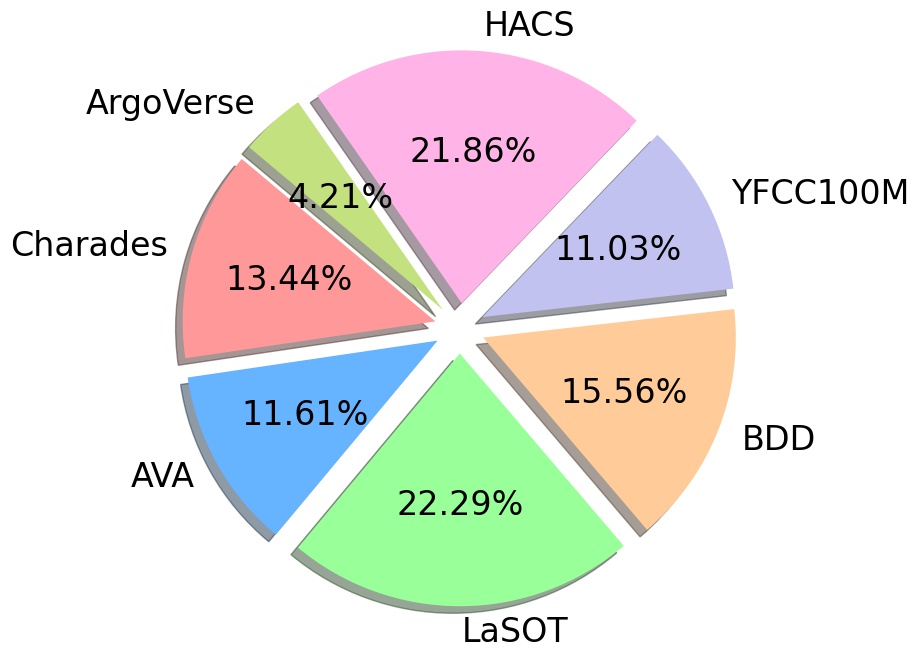}}
 \end{minipage} \hfill
 \begin{minipage}{.235\textwidth}
 \subcaptionbox{Interactivity types. \label{fig:stat_inter}}[1.0\textwidth]{\includegraphics[width=\textwidth]{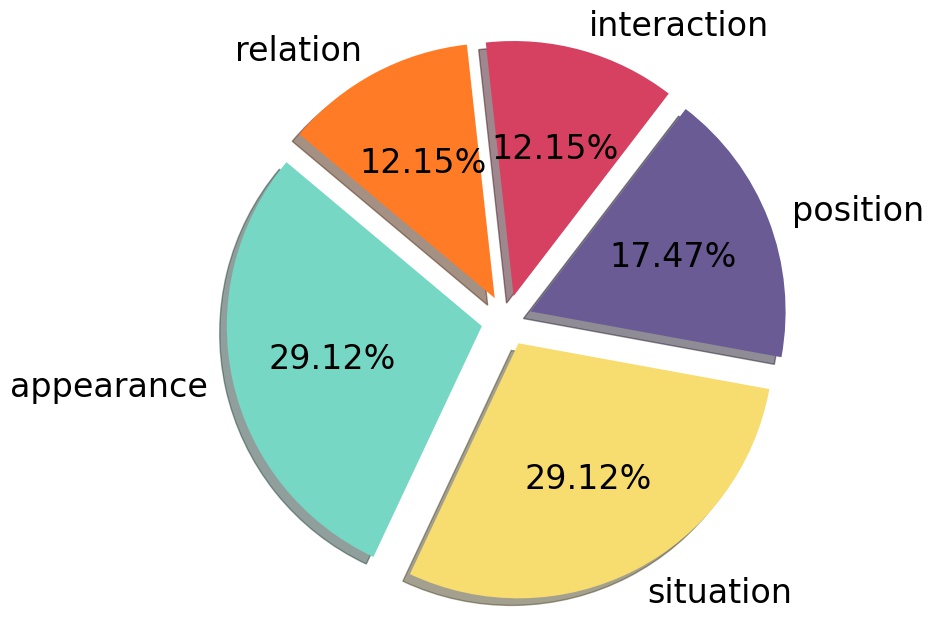}}
 \end{minipage}
 \caption{Statistics from the proposed \textit{ASPIRe} dataset.
 % The \textit{ASPIRe} dataset contains 400 third-person and ego-centric videos from diverse environments,as shown in (a). The statistics of object classes and relation classes are shown in (b) and (c).
 }
 \label{fig:data_stats}
 \vspace{-4mm}
\end{figure}
The \textit{ASPIRe} dataset is quantitatively analyzed in Table~\ref{tab:datasets} and visually represented in Fig.~\ref{fig:data_stats}. \textit{ASPIRe} contains 1,488 videos covering 833 object categories and 4,549 interactivities, including appearances, situations, positions, interactions, and relationships. The dataset is especially remarkable for its videos that depict a comprehensive and intricate variety of interactivities among subjects, with the number of appearances recorded at 722, situations at 2,902, positions at 130, interactions at 565, and relations at 230. Furthermore, the dataset features objects annotated with boxes and masks, amounting to 167,751 detail annotations.

We provide a detailed analysis of average occurrences within each video of the \textit{ASPIRe} dataset. On average, subjects are featured at 4.5 per video, showcasing diversity in the presence of objects. Both the frequency of appearances and situations remain steady at an average of 4.5 occurrences per video, suggesting a uniform representation of visual elements and their contextual narratives. Positions have a marginally lower average of 4.3 per video. Interactions and relationships averaged around 4.0 instances per video.

% \subsection{Benchmarking Protocols} 

\section{Methodology}

\subsection{Terminologies}\label{subsec:terms}
% \vspace{-1mm}
\begin{figure}[!t]
 \centering
 \includegraphics[width=\linewidth]{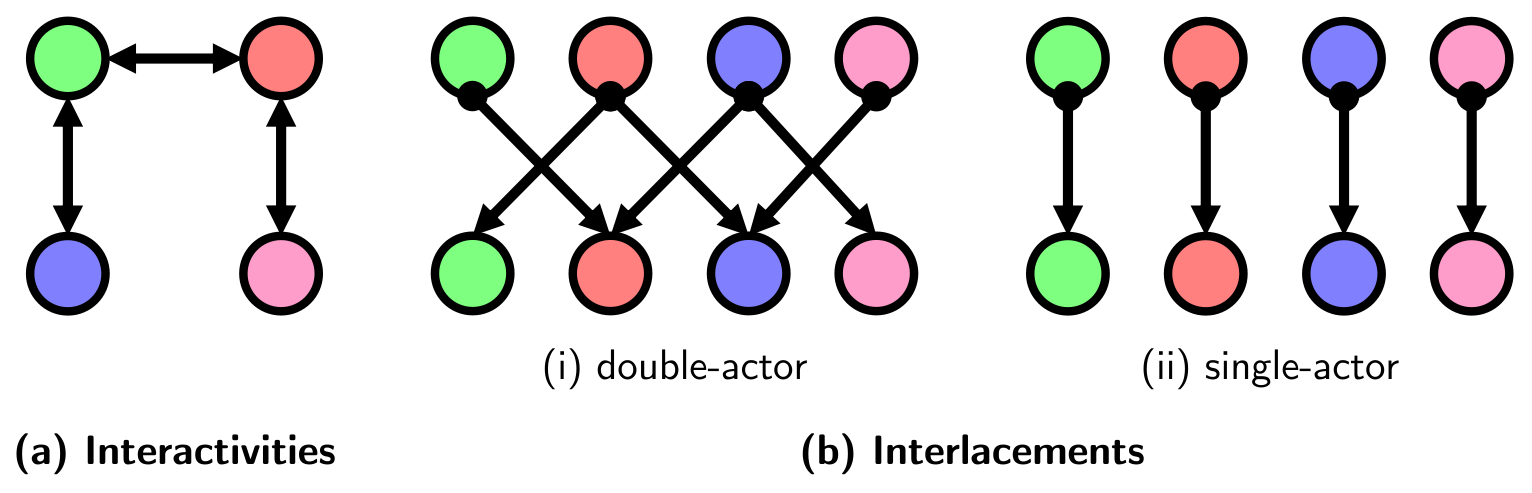}
 \caption{The terminologies used in our proposed \textit{ASPIRe} dataset and \textit{Hierarchical Interlacement Graph}.}
 \label{fig:terms}
 \vspace{-6mm}
\end{figure}

Fig.~\ref{fig:terms} illustrates our definitions for analyzing interactivities temporally. Fig.~\ref{fig:terms}a shows the original definition of interactivities within the subjects as annotated in our proposed \textit{ASPIRe} dataset. Interactivities refer to the relationship between subjects. Fig.~\ref{fig:terms}b illustrates a new term \textit{Interlacements}, which are interactivities that span across two or sets of nodes in time or frames. \textit{Interlacements} is our novel design representing how the interactivities evolve in our proposed HIG model, which will be present in the next Section~\ref{HIG}. Fig.~\ref{fig:terms}b has two parts, including \textit{double-actor} and \textit{single-actor} attribute interlacements. Fig.~\ref{fig:terms}b(i) defines double-actor attributes. double-actor attributess include \textit{position}, \textit{interaction}, and \textit{relation}, which are attributes that involve two subjects. Fig.~\ref{fig:terms}b(ii) defines single-actor attributes. Single-actor attributes include \textit{appearance} and \textit{situation}, attributes of individual subjects.

\subsection{Problem Formulation}
% \vspace{-1mm}
Given a video input $\in \mathbb{R} ^ {T \times H \times W \times 3}$ consisting of $ T $ frames and frame size of $ H \times W $, we identify a set of distinct subjects, represented as vertices in our graph, $V_t = \{S_i\}_t$ at a particular time $t$ and an interactivity set $I$ as in Eqn. \eqref{eqn:Eqn1}.
\begin{equation} \label{eqn:Eqn1}
\small
\begin{aligned}
I(S_i, S_j) = \Big\{\mathcal{A}(S_i), &\mathcal{S}(S_i),\\
& \mathcal{PO}(S_i, S_j), \mathcal{IN}(S_i, S_j), \mathcal{RE}(S_i, S_j)\Big\}
\end{aligned}
\end{equation}
It encapsulates all possible interactivities between subjects. Each element in $I$ provides a fine-grain classification of the interactivity types. These interactivities are appearance $\mathcal{A}(S_i)$, situation $\mathcal{S}(S_i)$ to express the single-actor attributes, and position $\mathcal{PO}(S_i, S_j)$, interaction $\mathcal{IN}(S_i, S_j)$ and relation $\mathcal{RE}(S_i, S_j)$ give the double-actor attributes, respectively. The primary objective is to construct a function $ f $. For each pair of subjects and each frame in the video, $ f $ identifies the most fitting interactivities from the set $ I $. This function is represented in Eqn. \eqref{eqn:problem}.
\begin{equation}\label{eqn:problem}
 \small
 f: V_t \times V_t \to I
\end{equation}
For every pair of objects drawn from $V_t$, the function $ f $ learns to predict an interactivity set $ I $, defining the Visual Interactivity Understanding task.

\section{Our Proposed Method}\label{HIG}

\begin{figure*}[th]
 \centering
 \includegraphics[width=.9\linewidth]{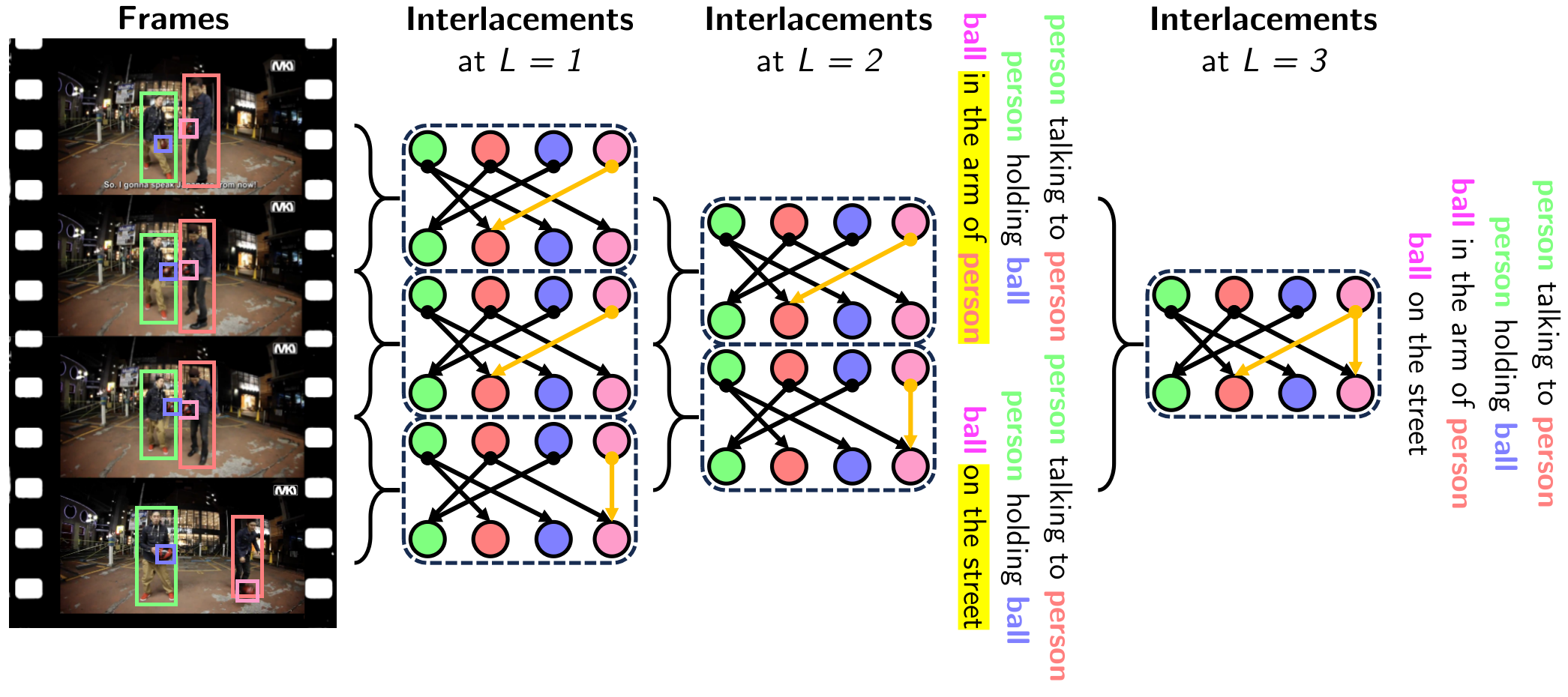}
 \caption{Our proposed \textit{Hierarchical Interlacement Graph}. The \colorbox{yellow!100}{highlighted attributes} denote the temporal changes in the graph. Then, all predicted interactivities are accumulated into the next hierarchy level. \textit{A higher-level graph cell covers a bigger portion of video frames}.}
 \label{fig:our_fw}
 \vspace{-4mm}
\end{figure*}

%\textcolor{red}{[@KL: Add one-two sentences here for introduction. Then connect to subsections.]}
% We propose the Hierarchical Interlacement Graph (HIG) model, a hierarchical architecture that captures the dynamic interplay of subjects within a video through a multi-level graph representation. Furthermore, we detail the training loss strategy, refining the model's ability to predict interactivity accurately.
Eqn.~\eqref{eqn:problem} is the primary objective in this problem. Our design of the graph structure, as in Fig.~\ref{fig:our_fw}, will be described below.

% \subsection{Learning Unified Interlacements}
\noindent \subsection{Hierarchical Interlacement Graph (HIG)}
% \vspace{-1mm}
\noindent\textbf{HIG} model is designed to capture the complex dynamics of object interactivity across both spatial and temporal dimensions~\cite{cetintas2023unifying}. It represents a video as a sequence of graphs $\{G_t(V_t, E_t)\}_{t=1}^{T}$ at the first layer, where each graph $G_t$ corresponds to a pair of frames. Here, $V_t$ denotes the set of nodes, and $E_t$ represents the set of edges at time $t$. As the model progresses through subsequent layers, it combines graphs from the previous layer to form new, more comprehensive graphs, culminating in a single graph cell at the highest level $L$, representing the entire video interlacement.

\noindent\textbf{HIG Blocks}. The HIG model consists of HIG blocks, each representing a distinct level of interactivity within the hierarchical structure. These blocks function consistently across all levels $l \in \{1, \dots, L\}$. At each level $l$, the model integrates graphs from the previous level to enhance the understanding of interactivity across spatial and temporal dimensions, as detailed in Algorithm~\ref{algm:hig_temporal}.

The feature representation $\mathcal{F}^{(l)}_t(S_i)$ is dynamically updated for every node $S_i$ at each level $l$ and time frame $t$. This update involves transformations and aggregations of information from the neighboring nodes of $S_i$. Each node $S_i$ in the graph encapsulates a feature set that evolves through the hierarchical levels, progressing horizontally across levels and vertically across time frames, starting from $t=1$ to $T_l = T - l + 1$ at each level. Specifically, at each level, the model transitions from processing a larger number of simpler graphs to fewer, more complex graphs. The feature representation $\mathcal{F}^{(l)}_t(S_i)$ at level $l$, with $l > 1$, is derived by aggregating transformed features of neighboring nodes from the previous level $l-1$ as shown in Eqn. \eqref{equa:layer}.
\begin{equation} 
    \small
    \mathcal{F}^{(l)}_t(S_i) = \sum_{S_j \in \mathcal{N}(S_i)} \mathcal{F}^{(l-1)}_t(S_j)
    \label{equa:layer}
\end{equation}
In Eqn.~\eqref{equa:layer}, the feature representation of a node at level $l$ is the sum of the transformed features of its neighboring nodes from the previous level. For each node $S_i$, the function $\mathcal{N}$ identifies a set of neighboring nodes that share similar attributes based on similarity scores. This procedure enhances the comprehensiveness of each node feature set as it ascends through the hierarchical layers.

\begin{algorithm}[!t]
 \footnotesize
 \caption{HIG Construction and Feature Embedding}
 \begin{itemize}
 \item \textbf{Input:} Frames as graphs $\{G_t(V_t, E_t)\}_{t=1}^{T}$; initial features $\mathcal{F}_t^{(0)}(S_i)$ for each node $S_i$; number of hierarchical levels $L$; weight matrices $\mathcal{W}^{(l)}_{ij}$ for all levels $l \in \{1, \ldots, L\}$ and node pairs $S_i, S_j \in V_t$.
 \item \textbf{Output:} $I(S_i, S_j)$
 \end{itemize}
 \begin{algorithmic}[1]
 \For{$l = 1$ to $L$}
 \State $T_l \leftarrow T - l + 1$  % Adjust time frame range for each level
 \For{$t = 1$ to $T_l$}
 \State $G_{l,t}(V_{l,t}, E_{l,t}) \leftarrow \mathrm{ConstructGraph}(G_t, l)$  % Construct graph for each level and time frame
 \For{$S_i \in V_{l,t}$}
 \State $m^{(l)}_t(S_i, S_j) \leftarrow \mathcal{W}^{(l)}_{ij} \cdot \mathcal{F}^{(l-1)}_t(S_j), \forall S_j \in \mathcal{N}(S_i)$  % Message computation
 \State $ \mathcal{F}^{(l)}_t(S_i) \leftarrow \sum_{t=1}^{T_l}\mathcal{F}^{(l-1)}_t(S_j), \forall S_j \in \mathcal{N}(S_i)$  % Feature update
 \EndFor
 \EndFor
 \EndFor
 \State $(V'_t, E'_t) \leftarrow (V'_{L,T_L}, E'_{L,T_L})$  % Final graph structure
 \State $\{\mathcal{F}'_t(S_i)\}_{S_i \in V'_t} \leftarrow \{\mathcal{F}'_{L,T_L}(S_i)\}_{S_i \in V'_{L,T_L}}$  % Final feature set
 % Derive Interactivity Feature:
 \For{$(S_i, S_j) \in V'_t \times V'_t$}
 \State $I(S_i, S_j) \leftarrow \mathcal{C}\left(m^{(L)}_{1}(S_i, S_j), \mathcal{F}^{(L)}_1(S_i)\right)$  % Interactivity feature computation
 \EndFor
 \end{algorithmic}
 \label{algm:hig_temporal}
\end{algorithm}

\noindent\textbf{Message-Passing Mechanism}. In our hierarchical design, nodes are interconnected through a message-passing mechanism. The message $m^{(l)}_t(S_i, S_j)$ at level $l$ and time $t$ is influenced by the weight matrix $\mathcal{W}^{(l)}_{ij}$ and the feature vector $\mathcal{F}^{(l-1)}_t(S_j)$ transmitted from $S_j$ to $S_i$. The message from node $S_j$ to $S_i$ is represented as in Eqn. \eqref{equa:mess}.
\begin{equation}
 \small
 m^{(l)}_t(S_i, S_j) = \mathcal{W}^{(l)}_{ij} \cdot \mathcal{F}^{(l-1)}_t(S_j)
 \label{equa:mess}
\end{equation}
In Eqn.~\eqref{equa:mess}, the message is a product of the weight matrix specific to that level and the feature vector of the sending node. The message $m^{(l)}_t(S_i, S_j)$ is transmitted from node $S_j$ to node $S_i$ shaped by the dimensions of the weight matrix $\mathcal{W}^{(l)}_{ij}$ and the feature vector $\mathcal{F}^{(l-1)}_t(S_j)$. The weight matrix $\mathcal{W}^{(l)}_{ij}$, critical at level $l$, typically has a shape of $(D_l \times D_{l-1})$, where $D_l$ denotes the feature dimension at level $l$ and $D_{l-1}$ represents the dimension at the preceding level $l-1$. Simultaneously, the feature vector of the node $S_j$ from the previous layer, denoted as $\mathcal{F}^{(l-1)}_t(S_j)$, is represented as a column vector with dimensions of $(D_{l-1} \times 1)$.

\noindent\textbf{Hierarchical Aggregation}. As the HIG model traverses its hierarchical structure, it progressively aggregates and refines node features from the initial to the final level. This transition involves combining and transforming node features, ensuring that the intricate details captured at lower levels are seamlessly integrated into the higher-level context. The process culminates at the highest level $L$, where the model consolidates all the refined features into a single graph cell at $t = 1$, as represented in Eqn. \eqref{equa:final}.
\begin{equation}
    \small
    \mathcal{F}^{(L)}_1(S_i) = \sum_{S_j \in \mathcal{N}(S_i)} \mathcal{F}^{(L-1)}_1(S_j)
    \label{equa:final}
\end{equation}
Eqn.~\eqref{equa:final} indicates the final feature representation $\mathcal{F}^{(L)}_1(S_i)$ at level $L$ is an aggregation of the transformed features of its neighboring nodes from the previous level. This final representation encapsulates the comprehensive interactivity information from all hierarchical levels.

\noindent\textbf{Interactivity Prediction}. For every pair of nodes $(S_i, S_j)$, the function $\mathcal{C}$ is employed to analyze their interactivity. This function considers both the message $m^{(L)}_{1}(S_i, S_j)$, which encapsulates the interactivity between the nodes, and the feature representation $\mathcal{F}^{(L)}_1(S_i)$, which reflects the features of the node $S_i$ at the highest hierarchical level. The prediction function is formulated as in Eqn. \eqref{equa:predict}.
\begin{equation}
\small
I(S_i, S_j) = \mathcal{C}\left(m^{(L)}_{1}(S_i, S_j), \mathcal{F}^{(L)}_1(S_i)\right)
\label{equa:predict}
\end{equation}
In Eqn.~\eqref{equa:predict}, $I(S_i, S_j)$ represents the predicted interactivities between nodes $S_i$ and $S_j$. The classification function $\mathcal{C}$ operates on the features and messages at the highest hierarchical level to produce a fine-grained classification on the edge connecting these nodes. The output of this function is represented in the set $I$, where each element provides a detailed classification of the five interactivity types, including appearance ($\mathcal{A}$), situation ($\mathcal{S}$), position ($\mathcal{PO}$), interaction ($\mathcal{IN}$), and relation ($\mathcal{RE}$).

% \noindent \textbf{Advantages.}
Designing a framework as our HIG model, involving data with varying subjects has distinct advantages. First, graphs are well-suited for the task, where the number of subjects can vary. Second, the message-passing mechanism allows interactivities to be exchanged between neighboring nodes.
% Those two advantages contribute to our \textit{flexibility} in defining both single-actor and double-actor attributes, and the number of actors can increase or decrease over time. 
Finally, HIG allows for a contextual understanding of where and when information occurs in the video, which is essential for tasks that require precise timestamps of events or actions.

% \noindent \textbf{Generalization.}

\subsection{Training Loss}
% \vspace{-1mm}
%\textcolor{red}{[@KL: Pls complete this section]}
The HIG model employs an integral training loss utilizing hierarchical weight sharing and sequential unfreezing techniques, with details provided in the following section.

\noindent \textbf{Sequential Training Strategy.}
The HIG framework employs a hierarchical weight-sharing strategy to enhance the efficiency of the training process. By sharing weights across different levels of the GNN hierarchy, the model takes advantage of a reduction in the total number of parameters, which operates as a regularizing mechanism to improve model generalization. In particular, training within the HIG framework is conducted through a sequential unfreezing strategy. Initially, the base level is activated, and subsequent levels are progressively unfrozen. This strategy allows the network to adapt to the feature embeddings  $\mathcal{F}^{(l)}_t(S_i)$, which are refined at each level $l$ and time step $t$.

At each level, the Focal Loss function~\cite{lin2017focal} is employed for edge classification, following~\cite{yang2022panoptic, yang2023panoptic, ji2020action}, as in Eqn. \eqref{equa:focal}. 

% \textcolor{red}{[@KL: Why focal loss?]}.
%
\begin{equation} \label{equa:focal}
\small
 \mathcal L(\mathcal{F}^{(l)}_t(S_i)) = -\alpha_t (1 - p_t(\mathcal{F}^{(l)}_t(S_i)))^\gamma \log(p_t(\mathcal{F}^{(l)}_t(S_i)))
\end{equation}
where $p_t$ measures the probability for the class, $\alpha_t$ is a weighting factor, and $\gamma$ is a parameter that adjusts the rate.

\noindent\textbf{Loss Aggregation.} The losses computed at each hierarchical level are aggregated to determine the total loss for the model as in Eqn. \eqref{equa:allloss}. This aggregation ensures that the training signal is comprehensive and encapsulates the learning objectives at each hierarchy level. The HIG framework promotes a nuanced training process, empowering the GNN to model the inherent hierarchical structures.
\begin{equation} \label{equa:allloss}
\small
 \mathcal L_{\text{total}} = \sum_{l} \mathcal L(\mathcal{F}^{(l)}_t(S_i))
\end{equation}

%\textcolor{red}{[@KL: Revise up to this point.]}

\section{Experiment Results}
% \vspace{-1mm}
\subsection{Implementation Details}
% \vspace{-1mm}
\noindent \textbf{Dataset.} The training set comprises 55K subjects and 197K interactivities across 500 videos. The validation set, which is used as the test set, comprises 988 videos with 113K subjects and 400 interactivities. In addition, we use PSG~\cite{yang2022panoptic} to evaluate our performance on the image data.
% \textcolor{red}{only describe the training and validation sets. How about the test set?}
% \textbf{ASPIRe}:

% \textbf{PSG}: 

\noindent \textbf{Model Configurations.}
This work uses the PyTorch framework and operates on 8$\times$ NVIDIA RTX A6000 GPUs. It utilizes a training batch size of 1 and employs the \texttt{AdamW} Optimizer, starting with an initial learning rate of 0.0001. We employ PyTorch Geometric~\cite{Fey/Lenssen/2019} for constructing graphs where nodes represent detections and edges signify potential interactivities. It integrates a ResNet-50~\cite{he2016deep} backbone trained with DETR~\cite{carion2020end}. 
% The framework has two pivotal classes: \texttt{Graph} and \texttt{HierarchicalGraph} from \texttt{torch\_geometric.data.Data}. The \texttt{Graph} class manages single-layer graphs using \texttt{torch\_scatter} while \texttt{HierarchicalGraph} is tailored for hierarchical scene data, handling multiple layers and aggregating node features. 
Our framework involves edge pruning using \texttt{scatter\_min} and \texttt{scatter\_max} for aggregating node features such as bounding box coordinates and track identification. Then, the framework calculates cosine similarity and selects the \textit{top-k} ($k = 12$) nearest neighbors.

\noindent \textbf{Metrics.}
Inspired by~\cite{shang2019annotating, yang2022panoptic, yang2023panoptic}, we calculate the recall metric for the Visual Interactivity Understanding task to predict a set of triplets that accurately describe the input video. The model predicts the category labels for the subject, object, and predicate within each triplet. Each triplet represents a distinct interactivity in the range time $t_1$ and $t_2$. Moreover, each triplet corresponds to a specific subject in single-actor scenarios and a pair of subjects in double-actor scenarios based on a predefined set. To this end, we leverage the standard metrics used in activity understanding, including $R@K$ and $mR@K$ utilized to evaluate the recall of top $K$ categories and their mean recall, respectively.

\subsection{Ablation Study}
% \vspace{-1mm} 
\begin{figure*}[t]
    \centering
    \includegraphics[width=.95\linewidth]{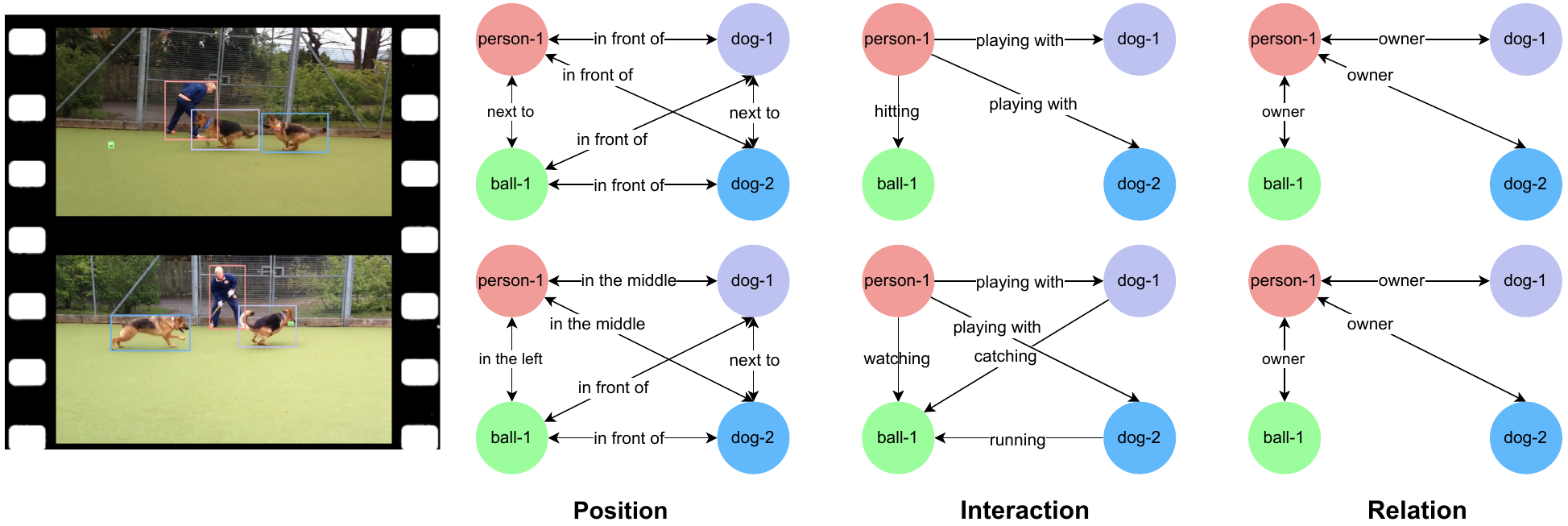}
    \caption{Qualitative results of position, interaction, and relation from scene graphs generated from the HIG model.
    % \textbf{Best viewed in color and zoom in}.
    }
    \label{fig:prediction}
    \vspace{-5mm}
\end{figure*}

\begin{table}[!t]
 \centering
 \caption{\footnotesize Comparison against baseline methods on single-actor attributes.}
 \label{tab:experiment_single_actor}
 \resizebox{\columnwidth}{!}{%
\begin{tabular}{c|crrr}
 \hline
 \textbf{Method} &
 \multicolumn{1}{c}{\textbf{Interlacement}} &
 \multicolumn{1}{c}{\textbf{R/mR@20}} &
 \multicolumn{1}{c}{\textbf{R/mR@50}} &
 \multicolumn{1}{c}{\textbf{R/mR@100}} \\ \hline
 \multirow{2}{*}{\textbf{Vanilla}} & Appearance &  10.88 / 0.09 & 12.19 / 0.09 & 14.16 / 0.08  \\
 & Situation &  2.87 / 0.02 & 5.29 / 0.03 & 9.05 / 0.03 \\ \hline
 \multirow{2}{*}{\textbf{Handcrafted}} & Appearance &  11.09 / 0.11 & 12.26 / 0.13 & 14.27 / 0.17  \\
 & Situation &  3.08 / 0.04 & 5.36 / 0.07 & 9.16 / 0.12  \\ \hline
 \multirow{2}{*}{\textbf{Convolution}} & Appearance &  11.32 / 0.11 & 12.28 / 0.25 & 14.32 / 0.22  \\
 & Situation &  3.31 / 0.04 & 5.38 /  0.19 & 9.21 / 0.17  \\ \hline
 \multirow{2}{*}{\textbf{Transformer}} & Appearance &  12.35 / \textbf{0.62} & 13.89 / \textbf{0.64} & 16.10 / \textbf{0.66}  \\
 & Situation &  4.54 / 0.55 & 6.99 / \textbf{0.58} & 10.99 /  0.61  \\ \hline
 \multirow{2}{*}{\textbf{HIG (Our)}} & Appearance & \textbf{15.02} / 0.60 & \textbf{18.60} / \textbf{0.64} & \textbf{20.11}/ 0.65 \\
 & Situation & \textbf{5.01} / \textbf{0.56} & \textbf{7.02} 
 / 0.55 & \textbf{12.01} / \textbf{0.63}\\ \hline
\end{tabular}
}
\vspace{-4mm}
\end{table}

\begin{table}[!t]
 \centering
 \caption{Comparison against previous methods on \textit{ASPIRe}.}
 \label{tab:sota}
 \resizebox{\columnwidth}{!}{%
\begin{tabular}{c|crrr}
 \hline
 \textbf{Method} &
 \multicolumn{1}{c}{\textbf{Interlacement}} &
 \multicolumn{1}{c}{\textbf{R/mR@20}} &
 \multicolumn{1}{c}{\textbf{R/mR@50}} &
 \multicolumn{1}{c}{\textbf{R/mR@100}} \\ \hline
 \multirow{2}{*}{\textbf{IMP~\cite{xu2017scene}}} 
 & Position & 9.70 / 0.49 & 9.70 / 0.49 & 9.70 / 0.49 \\
 & Interaction & 12.79 / 0.08 & 12.79 / 0.08 & 12.79 / 0.08 \\
 & Relation & 11.51 / 0.32 & 11.51 / 0.32 & 11.51 / 0.32  \\ \hline
 \multirow{2}{*}{\textbf{MOTIFS~\cite{zellers2018neural}}}  
 & Position & 6.89 / 0.48 & 8.49 / 0.38 & 8.70 / 0.40 \\
 & Interaction & 8.83 / 0.12 & 10.33 / 0.12 & 10.57 / 0.12  \\
 & Relation & 8.72 / 0.32 & 10.26 / 0.32 & 10.55 / 0.32  \\ \hline
 \multirow{2}{*}{\textbf{VCTree~\cite{tang2019learning}}} 
 & Position & 4.18 / 0.39  &  6.75 / 0.40  & 8.59 / 0.42 \\
 & Interaction & 6.23 / 0.10  &  9.58 / 0.10  & 11.63 / 0.10 \\
 & Relation & 6.51 / 0.27  & 9.82 / 0.28 & 11.51 / 0.28 \\ \hline
 \multirow{2}{*}{\textbf{GPSNet~\cite{lin2020gps}}} 
 & Position & 12.89 / \textbf{1.26} & 12.89 / 1.26 & 12.89 / \textbf{1.26} \\
 & Interaction & 10.89 / \textbf{0.11} & 10.89 / \textbf{0.12} & 10.89 / 0.12 \\
 & Relation & 9.87 / \textbf{0.35} & 9.87 / \textbf{0.35} & 9.87 / 0.35 \\ \hline
 \multirow{2}{*}{\textbf{HIG (Ours)}} 
 & Position & \textbf{13.02} / 0.09 & \textbf{24.52} / \textbf{1.33} & \textbf{42.33} / 1.12 \\
 & Interaction & \textbf{12.02} / \textbf{0.11} & \textbf{24.65} / \textbf{0.12} & \textbf{41.65} / \textbf{0.14} \\
 & Relation & \textbf{10.26} / 0.29 & \textbf{23.72} / 0.34 & 
 \textbf{41.47} / \textbf{0.39}\\ 
 \hline
 \end{tabular}%

 }\vspace{-6mm}
\end{table}

\begin{table}[!t]
 \centering
 \caption{\footnotesize Comparison against baseline methods on double-actor attributes.}
 \label{tab:experiment_double_actor}
 \resizebox{\columnwidth}{!}{%
\begin{tabular}{c|crrr}
 \hline
 \textbf{Method} &
 \multicolumn{1}{c}{\textbf{Interlacement}} &
 \multicolumn{1}{c}{\textbf{R/mR@20}} &
 \multicolumn{1}{c}{\textbf{R/mR@50}} &
 \multicolumn{1}{c}{\textbf{R/mR@100}} \\ \hline
 \multirow{3}{*}{\textbf{Vanilla}} & Position & 10.52 / 0.50 & 21.97 / 0.55 & 38.05 / 0.62 \\
 & Interaction & 10.16 / 0.12 & 22.35 / 0.13 & 39.91 / 0.14 \\
 & Relation & 9.71 / 0.32 & 21.96 / 0.36 & 39.11 / 0.40 \\ \hline
 \multirow{3}{*}{\textbf{Handcrafted}} & Position & 10.73 / 0.52 & 22.04 / 0.59 & 38.16 / 0.71 \\
 & Interaction & 10.37 / 0.14 & 22.42 / 0.17 & 40.02 / 0.23 \\
 & Relation & 9.92 / 0.34 & 22.03 / 0.40 & 39.22 / 0.49 \\ \hline
 \multirow{3}{*}{\textbf{Convolution}} & Position & 10.96 / 0.52 & 22.06 / 0.71 & 38.21 / 0.76 \\
 & Interaction & 10.60 / 0.14 & 22.44 / 0.29 & 40.07 / 0.28 \\
 & Relation & 10.15 / 0.34 & 22.05 / 0.52 & 39.27 / 0.54 \\ \hline
 \multirow{3}{*}{\textbf{Transformer}} & Position &   11.04 / \textbf{0.83} & 22.52 / 0.90 & 38.84 / 1.02   \\
 & Interaction &  10.68 / \textbf{0.45} & 22.90 / \textbf{0.48} & 40.70 / \textbf{0.52}  \\
 & Relation &  10.23 / \textbf{0.65} & 22.51 / \textbf{0.71} & 39.90 / \textbf{0.96} \\ \hline
 \multirow{3}{*}{\textbf{HIG (Ours)}} & Position & \textbf{13.02} / 0.09 & \textbf{24.52} / \textbf{1.33} & \textbf{42.33} / \textbf{1.12} \\
 & Interaction & \textbf{12.02} / 0.11 & \textbf{24.65} / 0.12 & \textbf{41.65} / 0.14 \\
 & Relation & \textbf{10.26} / 0.29 & \textbf{23.72} / 0.34 & \textbf{41.47} / 0.39\\ \hline
\end{tabular}
 }
 \vspace{-4mm}
\end{table}

\begin{table}[!t]
 \centering
 \caption{Comparison at different video sampling rates of our HIG.}
 \label{tab:s-rate}
 \resizebox{\columnwidth}{!}{%
 \begin{tabular}{c|crrrc}
 \hline
 \textbf{Sampling Rate} &
 \multicolumn{1}{c}{\textbf{Interlacement}} &
 \multicolumn{1}{c}{\textbf{R/mR@20}} &
 \multicolumn{1}{c}{\textbf{R/mR@50}} &
 \multicolumn{1}{c}{\textbf{R/mR@100}} &
\multicolumn{1}{c}{\textbf{FPS}} \\ \hline
 \multirow{5}{*}{\textbf{2 (Half)}} 
 & Appearance & 12.13 / 0.59	& 12.25 / 0.63	& 7.48 / 0.64 & \multirow{5}{*}{26.4} \\
 & Situation & 2.12 / 0.55 & 5.67 / 0.54	& 8.62 / 0.62 \\
 & Position & 10.13 / 0.08 & 18.17 / 1.32 & 29.7 / 1.11 & \\
 & Interaction & 9.13 / 0.10 & 18.30 / 0.11 & 29.02 / 0.13 & \\
 & Relation & 7.37 / 0.28 & 17.37 / 0.33	& 28.84 / 0.38 & \\ \hline
 \multirow{5}{*}{\textbf{1 (Full)}} 
 & Appearance & 15.02 / 0.60 & 18.60 / 0.64 & 20.11 / 0.65 & \multirow{5}{*}{24.2}\\
 & Situation & 5.01 / 0.56 & 7.02 / 0.55 & 12.01 / 0.63\\
 & Position & 13.02 / 0.09 & 24.52 / 1.33 & 42.33 / 1.12 & \\
 & Interaction & 12.02 / 0.11 & 24.65 / 0.12 & 41.65 / 0.14 & \\
& Relation & 10.26 / 0.29 & 23.72 / 0.34 & 41.47 / 0.39 & \\ 
 \hline
 \end{tabular}}
 \vspace{-4mm}
\end{table}

% \noindent \textbf{Generalization.}
% - generalization (prove the true generalization proof)

\noindent \textbf{Baseline Methods.} We re-implemented four baseline methods introduced in~\cite{yang2023panoptic} and presented in  Table~\ref{tab:experiment_single_actor} and Table~\ref{tab:experiment_double_actor} since the official implementation is unavailable. Table~\ref{tab:experiment_single_actor} compares all baseline methods and the HIG along single-actor attributes, and Table~\ref{tab:experiment_double_actor} compares double-actor attributes.
% \textcolor{red}{Lines 434 - 438, why is the evaluation not on each video frame? It seems that the current evaluation (indicated in Lines 434 - 438) is not well aligned with the task definition.}
HIG is designed to analyze videos through a hierarchical structure that progressively accumulates temporal information across multiple levels. Instead of getting results for each frame separately, as is done at level $l = 1$, we prefer the predictions made at higher levels, where the confidence score is greater $\geq 0.9$. A higher hierarchy level covers a more significant portion of the video frame, as in Fig.~\ref{fig:our_fw}. This approach \textit{effectively reduces noise and produces a higher recall rate}. In particular, the HIG method is better at recognizing single-actor attributes than other baselines, including Transformer, Convolution, Handcrafted, and Vanilla. Specifically, the HIG model is 2.67\% higher than the Transformer, the best method in baseline at $R@20$. HIG is also better for the double-actor attributes, especially in figuring out interactions and relations. It is 1.34\% higher than Transformers at $R@20$ when identifying interactions. We visualize keyframe predictions in a video, as shown in Fig.~\ref{fig:prediction}. 

\noindent \textbf{Video Sampling Rates.} Table~\ref{tab:s-rate} explores the influence of frame sampling rates on the performance of the HIG model in deployment. Our analysis focuses on evaluating the performance under a reduced number of frames. In the ASPIRe dataset, the testing set includes 988 videos, totaling 10,456,48 frames. We address the efficiency of the HIG model by halving the number of frames in each video. In particular, we discard one frame out of every two successive frames. Our experiment reveals a trade-off between recall score and inference time, where the HIG model experiences a decrease in recall performance but achieves a 2.2 FPS increase in inference speed.

\begin{table}[!t]
 \centering
 \caption{Comparison against previous methods on SGG task.}
 \label{tab:sgdet}
 \resizebox{\columnwidth}{!}{%
\begin{tabular}{c|crrr}
 \hline
 \textbf{Method} &
 \multicolumn{1}{c}{\textbf{Interlacement}} &
 \multicolumn{1}{c}{\textbf{R/mR@20}} &
 \multicolumn{1}{c}{\textbf{R/mR@50}} &
 \multicolumn{1}{c}{\textbf{R/mR@100}} \\ \hline
 \multirow{2}{*}{\textbf{IMP~\cite{xu2017scene}}} 
 & Position & 0.25 / 0.36 & 0.29 / 0.35 & 0.30 / 0.33 \\
 & Interaction & 0.71 / 0.13 & 0.98 / 0.12 & 1.15 / 0.13 \\
 & Relation & 0.80 / 0.26 & 0.81 / 0.25 & 0.84 / 0.24 \\ \hline
 \multirow{2}{*}{\textbf{MOTIFS~\cite{zellers2018neural}}}  
 & Position & 0.23 / 0.43 & 0.23 / 0.43 & 0.31 /0.38 \\
 & Interaction & 0.39 / 0.11 & 0.94 / 0.11 & 1.17 / 0.10 \\
 & Relation & 0.31 / 0.30 & 0.32 / 0.28 & 0.53 / 0.32 \\ \hline
 \multirow{2}{*}{\textbf{VCTree~\cite{tang2019learning}}} 
 & Position & 0.13 / 0.23 &  0.14 /  0.22 & 0.14 / 0.21 \\
 & Interaction & 0.55 / 0.07 &  0.65 / 0.08 & 0.72 / 0.08 \\
 & Relation & 0.39 / 0.18 & 0.39 / 0.20 & 0.43 / 0.21 \\ \hline
 \multirow{2}{*}{\textbf{GPSNet~\cite{lin2020gps}}} 
 & Position & 0.09 / 0.46 & 1.17 / 0.37 & 1.32 / 0.46 \\
 & Interaction & 0.99 / 0.09 & 1.02 / 0.09 & 1.11 / 0.09 \\
 & Relation & 0.14 / 0.23 & 0.16 / 0.13 & 0.29 / 0.23\\ \hline
 \multirow{2}{*}{\textbf{HIG (Ours)}} 
 & Position & \textbf{1.00} / \textbf{0.42} & \textbf{2.40} / \textbf{0.44} & \textbf{4.87} / \textbf{0.47} \\
 & Interaction & \textbf{1.30} / \textbf{0.09} & \textbf{3.45} / \textbf{0.11} & \textbf{6.93} / \textbf{0.12}  \\
 & Relation & \textbf{1.26} / \textbf{0.27} & \textbf{3.43} / \textbf{0.30} & \textbf{7.02} / \textbf{0.32} \\ \hline
\end{tabular}%
 }
 \vspace{-4mm}
\end{table}

\subsection{Comparison with State-of-the-Arts}
% \vspace{-2mm}
\textbf{Performance on \textit{ASPIRe}.} We provide the comparative analysis with SOTAs in Table~\ref{tab:sota}, including IMP~\cite{xu2017scene}, MOTIFS~\cite{zellers2018neural}, VCTree~\cite{tang2019learning}, and GPSNet~\cite{lin2020gps}. In the ASPIRe dataset, the HIG method shows impressive results in identifying the position on recall at different top $K$. In addition, the HIG model performs well on identifying relations when it is higher than 1.13\% at $R@20$ compared to GPSNet.

\noindent \textbf{Scene Graph Generation (SGG).} We extend the capability of the HIG model while incorporating image-based scene graph generation into the training process presented in Table~\ref{tab:sgdet}. Since the prior method was designed for interactions between pairs of subjects, we focus our comparison on the double-actor attributes. The HIG method demonstrates superior performance across all interlacement highlighting its advanced proficiency in attribute recognition within frame-based scene graph generation scenarios. Compared to the best-performing previous method, GPSNet, the HIG model achieves improvements of 3.55\%, 5.82\%, and 6.73\% at $R@100$ for position, interaction, and relation.

\noindent \textbf{Performance on PSG.} 
In addition to evaluating our method on a video dataset, we demonstrate its effectiveness on an image dataset by comparing it with state-of-the-art methods on the PSG dataset, as presented in Table \ref{tab:sota-psg}. When applied to the PSG dataset, the HIG model treats each image as a single-frame video, shifting its focus to \textit{spatial interactivity rather than temporal interactivity}. Although our model is primarily designed for video datasets, it achieves comparable results on the image dataset, with only a slight decrease at $R@20$ compared to state-of-the-art methods. Notably, the HIG model outperforms VCTree by 3.8\% in terms of $R@100$, highlighting the strength of the graph representation.

\begin{table}[!t]
 \centering
 \caption{Comparison against previous methods on PSG~\cite{yang2022panoptic}.}
 \label{tab:sota-psg}
 \resizebox{\columnwidth}{!}{%
 \begin{tabular}{c|crrr}
 \hline
 \textbf{Method} &
 % \multicolumn{1}{c}{\textbf{Interlacement}} &
 \multicolumn{1}{c}{\textbf{R/mR@20}} &
 \multicolumn{1}{c}{\textbf{R/mR@50}} &
 \multicolumn{1}{c}{\textbf{R/mR@100}} \\ \hline
 \multirow{1}{*}{\textbf{IMP~\cite{xu2017scene}}} 
 & 16.5 / 6.52 & 18.2 / 7.05 & 18.6 / 7.23 \\ \hline
 \multirow{1}{*}{\textbf{MOTIFS~\cite{zellers2018neural}}}  
  &  20.0 / 9.10 & 21.7 / 9.57 & 22.0 / 9.69 \\ \hline
 \multirow{1}{*}{\textbf{VCTree~\cite{tang2019learning}}} 
  &  \textbf{20.6} / \textbf{9.70} & 22.1 / 10.2 & 22.5 / 10.2 \\ \hline
 \multirow{1}{*}{\textbf{GPSNet~\cite{lin2020gps}}} 
  & 17.8 / 7.03 & 19.6 / 7.49 & 20.1 / 7.67 \\ \hline
\multirow{1}{*}{\textbf{PSGFormer~\cite{yang2022panoptic}}} & 18.6 / 16.7 & 20.4 / \textbf{19.3} & 20.7 / \textbf{19.7}\\ \hline
 \multirow{1}{*}{\textbf{HIG (Ours)}} 
  & 19.4 / 6.42 & \textbf{22.3} / 8.13  & \textbf{26.3} / 9.70  \\ \hline
 \end{tabular}%
 }
 \vspace{-4mm}
\end{table}

\section{Conclusion}
% \vspace{-2mm}
% \textcolor{red}{one conclusion paragraph}
We addressed the Visual Interactivity Understanding problem by introducing the \textit{ASPIRe} dataset and the \textit{Hierarchical Interlacement Graph}. APSIRe established a new benchmark with its extensive predicate types offering nuanced interactivity perspectives. Meanwhile, HIG provides a unified hierarchical structure for capturing complex video interlacements, demonstrating scalability and flexibility in handling five interactivity types. Additionally, we provided extensive experiments showcasing the efficiency of HIG and achieving state-of-the-art results in both video and image datasets. 

\noindent \textbf{Limitations.} While the HIG approach significantly advanced the understanding of interactivities, it faced certain limitations. 
Computing possible interlacements became a computational bottleneck, potentially hindering real-time applications. Also, the framework faced challenges in handling long-duration videos, where the continual learning of new interactivities could lead to the decay of previously acquired knowledge. As the HIG model was tailored for video datasets, its image-based performance might not be optimal.

% \newpage

\noindent \textbf{Acknowledgment.} This work is partly supported by NSF Data Science and Data Analytics that are Robust and Trusted (DART), NSF SBIR Phase 2, and Arkansas Biosciences Institute (ABI) grants. We also acknowledge the Arkansas High-Performance Computing Center for providing GPUs.

 %%%%%%%%% REFERENCES
 {\small
 \bibliographystyle{ieee_fullname}
 \bibliography{egbib}
 }

\end{document}